# Implementation Of Wildlife Observation System

Prof. Neethu K N
*Department of Electronics and Communication Engineering*
REVA University
Bengaluru, Karnataka
kn.neethu@reva.edu.in

Rakshitha Y Nayak
*Department of Electronics and Communication Engineering*
REVA University
Bengaluru, Karnataka
rakshithaynayak@gmail.com

Rashmi
*Department of Electronics and Communication Engineering*
REVA University
Bengaluru, Karnataka
rashmihiregoudar23@gmail.com

Meghana S
*Department of Electronics and Communication Engineering*
REVA University
Bengaluru, Karnataka
meghanas0529@gmail.com

*Abstract*—By entering the habitats of wild animals, wildlife watchers can engage closely with them. There are some wild animals that are not always safe to approach. Therefore, we suggest this system for observing wildlife. Android phones can be used by users to see live events. Wildlife observers can thus get a close-up view of wild animals by employing this robotic vehicle. The commands are delivered to the system via a Wi-Fi module. As we developed the technology to enable our robot to deal with the challenges of maintaining continuous surveillance of a target, we found that our robot needed to be able to move silently and purposefully when monitoring a natural target without being noticed. After processing the data, the computer sends commands to the motors to turn on. The driver motors, which deliver the essential signal outputs to drive the vehicle movement, are now in charge of driving the motors.

## I. INTRODUCTION

Today's wildlife is at risk from poaching and animal smuggling, which has resulted in the endangerment of most species. Therefore, we are here with a new concept i.e., Wildlife Observation Robot. The robot includes a rotating 360-degree night vision camera. This enables the user to wirelessly drive the robotic vehicle and get the necessary angle. It is possible to capture video or take pictures that can be afterwards reviewed. We may study the daily activities of wildlife with our observation robot. Furthermore, we can use this ground-breaking technology to capture up-close images of wild animals. A Raspberry Pi-3 is located in this module, and it is connected through WiFi to either an Android device or a laptop. The Raspberry Pi circuit put on the robot will retrieve the commands the user enters. In order to control the robot, the Raspberry Pi-3 analyses this data and sends signals to the motor drivers. Now, the driver motors drive the vehicle motors in turn by transmitting the necessary signal outputs. So, with the help of the Raspberry Pi-3 circuit, this wildlife observation robot with night vision capability system enables getting a closer view of wildlife.

## II. LITERATURE SURVEY

There has been research about Wildlife observation system in the past few years. Some of the papers mentioned below are the ones which we have referred.

"Android Controlled Wildlife Observation Robot" In this paper they employed a transmitter and receiver equipment for this essay. The transmitter unit is made up of a battery, an 8051 microcontroller for controlling the robot's motor and camera angle, a Bluetooth module for communicating with an Android device, an IC driver for the motor, a night vision wireless camera, and a servo motor for adjusting the camera's angle. The antenna for signal reception and the television for display make up the receiver device. As a result, this machine was created to keep an eye on wildlife without endangering people's lives because it is simple to see animals from a distance. [1].

"Android Controlled Robot for surveillance", In this work, a motor driver (L293D), a DC motor, an Arduino, the Arduino IDE, and a Bluetooth module (HC05) were utilised. The Arduino Uno receives the data from the Android app first as an input from the Bluetooth module. The Arduino Uno is a controller that manages signals and carries out specified tasks; it knows which signals should be sent to the motor driver so that it can move in a specific sequence. With the aid of the Bluetooth module, users instruct the microcontroller via the app. The Arduino then controls the motor driver, supporting the DC motors and turning on the high signal at particular motor pins. An ultrasonic sensor measures the obstacle's distance from the robot and outputs the determined distance as a serial output on the app screen. The robot could then be moved in any direction wanted in this manner. [2].

"IoT based robot control using smartphone", In this paper they have used an ATMEGA16, AVR studio, DipTrace, a Wi-Fi module, and a blynk programme have all been used in this study. First, in the project, a Wi-Fi connection was made between a smartphone and the internal Wi-Fi module of the robot. Once the ATMEGA16 microcontroller has received commands from the smartphone through programming, the L293D motor driver IC drives the DC motors in accordance with the instructions. Real-time videos of faraway locations (where humans are unable to travel) can be recorded with the use of a camera. The operator will then receive the footage for further action. [3].

"Wi-Fi Surveillance Robot Using Raspberry Pi" In this study, researchers used Wi-Fi to remotely control a robot using a smartphone or laptop and transmit live video from the robot for surveillance purposes. This video was retrieved from the remote device's web browser while the device was being used. DC motors were employed to control the robotic movement, and a stepper motor was used to move the camera. Wi-Fi and the internet are used to transport the processed video from the Raspberry Pi B+ to the user's PC for viewing. [4].

"Remotely Controlled Autonomous Robot using Android Application", In this study, a Raspberry Pi 3 was employed, which was connected to a variety of sensors, including an accelerometer and an ultrasonic sensor. To input commands to control the robot remotely, an Android smartphone is used. An Android application sends commands, and a Raspberry Pi 3 receives them. The Raspberry Pi instructs the robot to carry out the necessary action in accordance with the specified directions. Using an ultrasonic sensor, the object's presence is detected. The HC-SR04 ultrasonic sensor is used to measure distance, while the ADXL 345 accelerometer sensor is used to assess direction. The Raspberry Pi is provided the signal that was acquired from the sensor. To capture the image of the object, a Sony IMX219 Camera module with an 8-megapixel resolution is used. Both recording videos and taking pictures are done with the camera module. Using the camera CSI connector that is built into the board, this camera is connected directly to the Pi. Using the USB connection on the Raspberry Pi, another webcam can be connected. The robot can be moved in the desired direction using DC motors. The camera is moved by a servo motor. Speakers are used to transmit voice commands or prerecorded audio music. By employing speakers, parents can speak to their infant and the infant would be able to hear them. By employing speakers, parents can speak to their infant and the infant would be able to hear them. Through the camera on an android device, parents can watch their child from a distance and communicate with them via voice message using the android app. [5].

"Energy-Efficient Visual Eyes System for Wildlife" Since the in-situ situation could be quickly understood, wireless real-time surveillance systems are very helpful for monitoring wildlife ecology. The primary restriction in untamed areas is the availability of power sources. As a result, they've integrated an MSP430 micro controller, an FPGA circuit, a solar-powered battery, a digital camera, and a GPRS (General Packet Radio Service) modem in this work to create a low-power picture transmission system (dubbed energy-efficient visual eyeballs system). The proposed system is built with a packet transmission technique capable of improving pixel validity while significantly reducing power consumption, making it a monitoring system appropriate for uninhabited islands. Practically, the planned system is installed on a savage island in Taiwan close to the East China Sea, which serves as a habitat for migrating seabirds. The experimental findings demonstrate the system's viability for ecological monitoring and its ability to efficiently preserve the images' completeness while being energy-efficient. [6].

III. METHODOLOGY

*A. System Overview*

In this system, one can control the camera angle remotely i.e. mobile with the use of Wi-Fi and also one can get the live streaming of video from the robot for the purpose of observation, this video is obtained on android app of the remote device from where we are operating the camera. DC motors are being used for the movement of robotic wheels in the forward direction and servo motor is used for turning the robot to left and right directions. A Raspberry Pi-3 is used for controlling the movement of the robotic vehicle using the motor driver circuit. In addition to this there is an ultrasonic sensor used to detect the obstacle infront of the robotic vehicle . when the ultrasonic sensor detects the obstacle infront of the vehicle it stops the movement of the vehicle then the robotic vehicle turns left and right based on the programming in the raspberry pi.

The block diagram of Implementation of wildlife observation system is shown below:

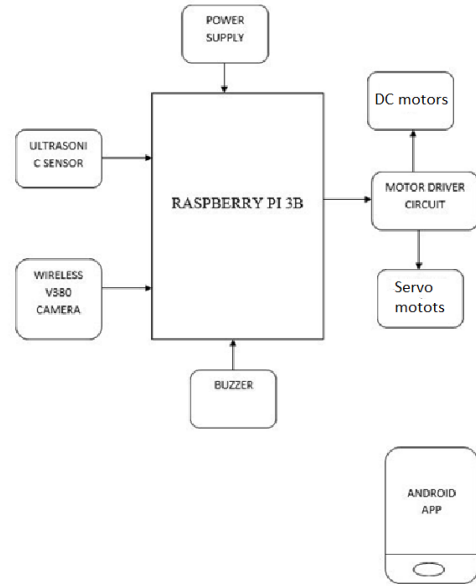

Figure 1: Block diagram Implementation of wildlife observation system

The circuit diagram of wildlife observation system is shown below:

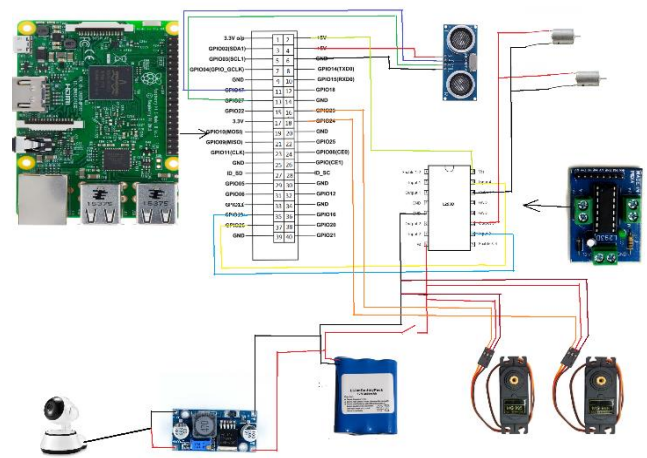

*1) RASPBERRY PI 3B.*

Raspberry pi is used for programming the robot to control the motors to move forward and stop when an obstacle is detected and turn left and right and continue moving.

*2) LM2596.*
This is a buck converter that is used to step down the voltage from the battery and give the output voltage to the camera.

*3) ULTRASONIC SENSOR*
Ultrasonic sensor is used to detect the obstacle in front of the robotic vehicle. Here we have programmed in such a way that when there is an obstacle the robot stops moving.

*4) DC MOTORS*
DC motors are used for the movements of the back wheels of the robotic vehicle.

*5) Servo MOTORS*
Servo motors are used for controlling the movement of the front wheels to turn left or right.

*6) MOTOR DRIVER CIRCUIT*
This motor driver circuit is used for controlling the movement of the motors that helps in the movement of the vehicle. The motor driver circuit used in this system is L293D.

*7) V380 CAMERA*
This camera is used for recording the video of the surrounding and sending the video to the android app for monitoring purpose.

**WILDLIFE OBSERVATION SYSTEM**

*B. System Flowchart*

Flowchart showing how the wildlife observation system operates. By providing the necessary electricity, the Raspberry Pi, Ultrasonic sensor, Wireless camera, and the DC motors are initialised. Once the switch is turned on, the user will be able to control the camera angle in the remote end and the robotic body will continue to move forward. The ultrasonic sensor stops the robot when it encounters an obstruction and directs it to turn left or right in accordance with the programme.

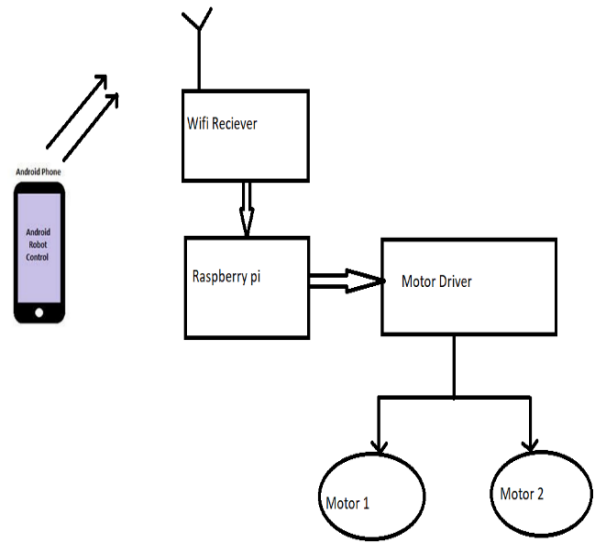

### IV. RESULTS

In this project, we hope to show that it is possible to capture close-ups of wild animals from any angle without endangering wildlife or people, and that these images can be used to track the animals and benefit researchers.

The following results were obtained:

*A. Hardware*

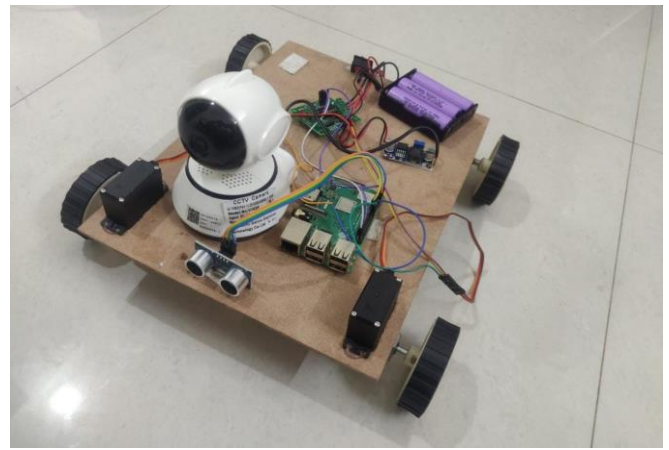

### V. CONCLUSION

The study concludes that using an Android-controlled wildlife observation robot will enable us to circumvent a number of challenges associated with wildlife observation. It will be simpler for observers and analysts to properly investigate the habitats of wildlife thanks to this new amalgamation of well-known technology. It is possible to save or give animals with a slimmer chance of surviving. Additionally, it will lessen the risk that people pose to themselves when they manually try to see or film dangerous animals. This machine serves as a companion for individuals. Because of how easily a layperson may use it, it can be beneficial. Additionally, the majority of those who need it can afford it.